\documentclass{article}

\usepackage[preprint]{neurips_2023}

\usepackage[utf8]{inputenc} 
\usepackage[T1]{fontenc}    
\usepackage{hyperref}       
\usepackage{url}            
\usepackage{booktabs}       
\usepackage{amsfonts}       
\usepackage{nicefrac}       
\usepackage{graphicx}
\usepackage{amsmath}
\usepackage{natbib}

\graphicspath{ {images/} }

\newcommand{\etal}{\textit{et al}.}

\title{Estimating Uncertainty with Implicit Quantile Network}
\author{
  Yi Hung Lim\\
  Rensselaer Polytechnic Institute\\
  110, 8th Street, Troy, NY\\
  \texttt{limy4@rpi.edu}\\
}

\date{\today}

\begin{document}

\maketitle

\begin{abstract}
  Uncertainty quantification is an important part of many performance critical applications. This paper provides a simple alternative to existing approaches such as ensemble learning and bayesian neural networks. By directly modeling the loss distribution with an Implicit Quantile Network, we get an estimate of how uncertain the model is of its predictions. For experiments with MNIST and CIFAR datasets, the mean of the estimated loss distribution is 2x higher for incorrect predictions. When data with high estimated uncertainty is removed from the test dataset, the accuracy of the model goes up as much as $10\%$. This method is simple to implement while offering important information to applications where the user has to know when the model could be wrong (e.g. deep learning for healthcare).
\end{abstract}

\section{Introduction}

The revolution in deep learning has led to many profound consequences across different fields. As deep learning models become more and more integrated into industry, uncertainty quantification becomes more and more important to ensure the safety of the users and the performance of the software. Many prior work has tried to tackle this problem by changing the main model architecture itself, leading to variants of neural networks such as bayesian neural networks. However, these approaches still fall short of ensemble models and dropout \citep{srivastava2014dropout, gal2016dropout}, and sometimes even decrease in accuracy compared to vanilla models. \citep{ovadia2019can} We propose a simple add-on to any deep learning model that would benefit from uncertainty quantification. By repurposing Implicit Quantile Network to predict the loss distribution of the prediction model on the training set, we can get an estimate of the uncertainty of the model on the test set. This approach does not require any architecture change to the vanilla model and does not require as much compute as ensemble models to train many independent copies of the same model.

\section{Background / Related Work}

\subsection{Implicit Quantile Networks for Distributional Reinforcement Learning}

Deep Q Network agents are a class of reinforcement learning algorithms that uses Q learning to try to learn a policy that would maximize the expected return \citep{mnih2013playing}. However, these agents often output a single scalar estimate of the return value, which does not take into account the randomness of its environments. Implicit Quantile Network was proposed by Dabney \etal \citep{dabney2018implicit} to introduce a new DQN-like agent that approximates the entire distribution of the return value instead of regressing the expected value. By randomly sampling $\tau \sim U([0,1])$, we can approximate a value for each quantile (aka percentile) of the distribution. During inference, batches of $\tau$ are sampled to approximate the true distribution. IQN outperforms DQN by a wide margin both in terms of sample efficiency and final performance. A proposed explanation for this gap is that IQN does not suffer from noisy gradients where the scalar target varies significantly due to the inherent randomness of its environment. IQN uses a quantile regression loss to try to approximate a better distribution target given its current predicted distribution.


\subsection{Ensemble Models}

Ensemble Models \citep{parker2013ensemble} is an approach to modeling uncertainty by training many independent copies of the same model. By doing so, any disagreement among these models can be labeled as uncertainty. By using significantly more compute, these models are robust when given data that are out of distribution because predictions from independent copies will disagree with each other when their predictions are wrong or if the data is out-of-distribution. Empirically, this method yields the best results in terms of accuracy and uncertainty quantification.

\subsection{Dropout for Uncertainty Estimation}

Another approach to estimate uncertainty is to use dropout \citep{srivastava2014dropout, gal2016dropout}. If the model is certain of its prediction, then dropping some units along with their connections will still result in the same prediction. However, if the predictions vary, then the difference in output can be interpreted as uncertainty. Some approaches try to incorporate dropout at training time while other work try to mimic an ensemble of models by only incorporating dropout at test time.

\subsection{Bayesian Neural Networks}

Bayesian Neural Networks attempt to quantify uncertainty by learning a posterior distribution of the weights. Therefore, each weight sample produces different outputs to form a distribution which captures the modes of the output. Although there are theoretical benefits for using BNN, it can be computationally expensive and sometimes fall short in accuracy.

\section{Preliminaries}

We briefly review the formulation for training IQN. Instead of the computing $Q(x, a)$, we define $Z(x, a)$ as the distribution of all possible returns. Then, we can get $Q(s, a)$ from $Z(s, a)$ given $Q_\beta(x, a) := \mathbb{E}_{\tau \sim U([0,1])} \left[ Z_{\beta(\tau)}(x, a) \right]$. The policy would simply be to maximize the expectation where $\pi_\beta(x) = argmax_{a \in A} Q_\beta(x, a)$. During training, we randomly sample $\tau \sim U([0,1])$ and use the quantile regression loss \citep{huber1992robust} to minimize the TD-errors. Formally,

\begin{equation}\label{eqn:sampledTD}
    \delta^{\tau,\tau'}_t = r_t + \gamma Z_{\tau'}(x_{t+1}, \pi_\beta(x_{t+1})) - Z_{\tau}(x_t, a_t).
\end{equation}

Then, we can train the quantile estimates with threshold $\kappa$. (Alternatively, we can use the mean squared error loss instead of the huber loss for quantile regression)
    
\begin{align*}
    \rho^\kappa_\tau(\delta_{ij}) = |\tau - \mathbb{I}{\{ \delta_{ij} < 0 \}}| \frac{\kappa(\delta_{ij})}{\kappa},\ \quad \text{with the Huber loss as}\\
    \kappa(\delta_{ij}) = \begin{cases}
        \frac{1}{2} \delta_{ij}^2,\quad \ &\text{if } |\delta_{ij}| \le \kappa\\
        \kappa (|\delta_{ij}| - \frac{1}{2}\kappa),\quad \ &\text{otherwise}
    \end{cases}
\end{align*}

For two samples $\tau, \tau' \sim U([0,1])$, and policy $\pi_\beta$, the sampled temporal difference (TD) error at step $t$ is \\
 
\begin{equation}\label{eqn:iqn_loss}
    \mathcal{L}(x_t, a_t, r_t, x_{t+1}) = \frac{1}{N'} \sum_{i=1}^{N} \sum_{j=1}^{N'} \rho_{\tau_i}^\kappa \left( \delta_t^{\tau_i, \tau_j'} \right)
\end{equation}

where $N$ and $N'$ denote the respective number of iid samples $\tau_i, \tau_j' \sim U([0,1])$ used to estimate the loss. Intuitively, for a given $\tau_i=0.75$, the pareto optimal point would be at the 75th percentile of all the observed values where the errors from the left and right hold equal weight, since negative errors are weighted 0.25 and positive errors are weighted 0.75.

\section{Modeling Loss with Implicit Quantile Network}

In the reinforcement learning setting, Implicit Quantile Network takes in a state and sample many $\tau \sim U([0,1])$ to output an approximated distribution of Q values. For our work, we aim to quantify uncertainty using IQN for the supervised learning setting. During training, we sample $\tau_i$ to approximate the distribution of the loss scalar, although we set the $N=64$ and $N'=1$, the model can still approximate the loss distribution over many iterations as the quantiles are trained to converge to its respective values.

Existing work has only used IQN to model rewards in reinforcement learning settings. This paper shows that we can directly predict the loss of the supervised learning model to get an estimated error which can then be used to quantify how certain a model is about its prediction. This approach requires additional compute by training a separate neural network that approximates the loss distribution of the original model after training.

Before regressing the loss, we train the main model on the dataset as usual, the trained weights are then transferred to the IQN after training which is then re-trained to predict the loss distribution on the training set.

For all experiments, N taus are sampled at each iteration with each representing a percentile of the distribution. The IQN then outputs a predicted loss distribution based on the tau values which is then trained using the quantile regression loss. We hypothesize that estimating loss quantifies both aleatoric uncertainty (data uncertainty) and epistemic uncertainty (model uncertainty) since it is both a measure of poorly labeled data and suboptimal weights.

\section{Experiments}

This simple approach is tested on common image classification benchmarks including MNIST and CIFAR \citep{deng2012mnist, krizhevsky2009learning} to determine whether IQN can correctly predict which images the model is more likely to get wrong. Since the increased accuracy could be a direct result of estimating the loss, experiments are also done for models where we only predict a scalar estimate to benchmark against the gains of predicting a distribution. Code is released on Github. \texttt{https://github.com/YHL04/confidenceiqn}

\subsection{Experiment Setting}

All experiments are done with a Convolutional Neural Network as backbone for all models. All benchmarks are trained in 20 epochs. We use the Adadelta optimizer with a learning rate of 1.0 and step learning rate $\gamma$ of 0.7. In addition, IQN is also benchmarked on images that are completely pitch black (hold no information) to see whether the estimated distribution agree that it should be very uncertain of its prediction.

\subsection{Results}

Below are the computed statistics for the estimated loss distribution on the datasets after training. For MNIST examples, the IQN has a predicted mean 10x higher than the dataset mean for incorrectly predicted labels while the scalar model is unable to distinguish as clearly with only a slight increase in estimated loss.

\begin{center}
\begin{tabular}{  c c c c  }

\hline
Scalar Model & MNIST & CIFAR10 & CIFAR100 \\
\hline
Mean & 0.003 & 0.301 & 1.884 \\
Std & 0.001 & 0.300 & 0.863 \\
Incorrect & \textbf{0.004} & 0.407 & 2.075 \\
Correct & 0.003 & 0.260 & 1.581 \\
Zeros & \textbf{0.004} & \textbf{1.205} & \textbf{2.619} \\
\hline

\end{tabular}
\end{center}

\begin{center}
\begin{tabular}{  c c c c  } 
\hline
IQN Model & MNIST & CIFAR10 & CIFAR100 \\
\hline
Mean & 0.003 & 0.315 & 1.760 \\
Std & 0.019 & 0.314 & 0.893 \\
Incorrect & \textbf{0.041} & \textbf{0.458} & \textbf{2.002} \\
Correct & 0.003 & 0.261 & 1.396 \\
Zeros & 0.039 & 1.144 & 2.327 \\
\hline

\end{tabular}
\end{center}

Since labels that are predicted incorrectly are more likely to have a higher estimated loss distribution, by setting a threshold for the estimated loss, we can then remove the predictions that are N standard deviation above the mean. This leads to higher accuracy on the test set. We use dropout of 0.25 after convolutional layers and dropout of 0.50 after the first linear layer unless stated otherwise.
\\

\begin{center}
\begin{tabular}{  c c c c  }

\hline
Original Model Accuracy & MNIST & CIFAR10 & CIFAR100 \\
\hline
No Dropout & 99.09 & 70.64 & 38.16 \\
Scalar & 99.19 & 72.55 & 38.43 \\
IQN & 99.25 & 73.22 & 38.72 \\
\hline

\end{tabular}
\end{center}

\begin{center}
\begin{tabular}{  c c c c  }

\hline
Model Accuracy (N = 0) & MNIST & CIFAR10 & CIFAR100 \\
\hline
Scalar & \textbf{99.73} & 81.29 & 50.87 \\
IQN & 99.45 & \textbf{81.73} & \textbf{51.68} \\
\hline

\end{tabular}
\end{center}

\begin{center}
\begin{tabular}{  c c c c  }

\hline
Model Accuracy (N = 0.5) & MNIST & CIFAR10 & CIFAR100 \\
\hline
Scalar & 99.19 & 77.90 & 43.29 \\
IQN & \textbf{99.38} & \textbf{78.51} & \textbf{46.54} \\
\hline

\end{tabular}
\end{center}

\begin{center}
\begin{tabular}{  c c c c  }

\hline
Model Accuracy (N = 1.0) & MNIST & CIFAR10 & CIFAR100 \\
\hline
Scalar & 99.19 & 74.29 & 40.52 \\
IQN & \textbf{99.37} & \textbf{76.54} & \textbf{42.32} \\
\hline

\end{tabular}
\end{center}

Results show that the IQN model consistently outperforms the baseline (regressing the loss with a single output). Note that if we remove all labels with estimated loss bigger than mean (N = 0), we get a 13\% improvement in accuracy for CIFAR100. 

\subsection{Distributions}

The MNIST and CIFAR100 samples and its predicted distributions are shown below. According to the results, handwritten digits that are more ambiguous have a substantially higher loss distribution. The red line is the mean over all the samples. Each sample uses 10000 tau samples for the distribution. From the visualization of the distribution, we see that the pitch black image has very high estimated loss distribution while clearer handwritten digits have below average distributions. Even though the pitch black image is OOD, it is still able to consistently predict a high distribution. An advantage of IQN is that is it able to capture all the modes of the distribution to account for different cases (different probability densities) instead of simply regressing the expected value.









\includegraphics[width=6.40cm, height=4.80cm]{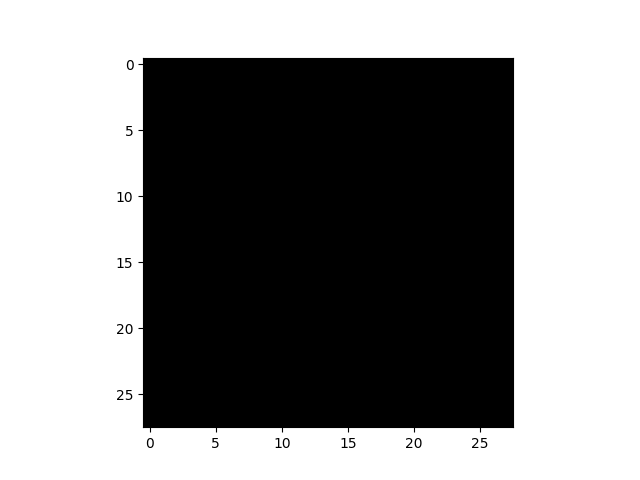}
\includegraphics[width=6.40cm, height=4.80cm]{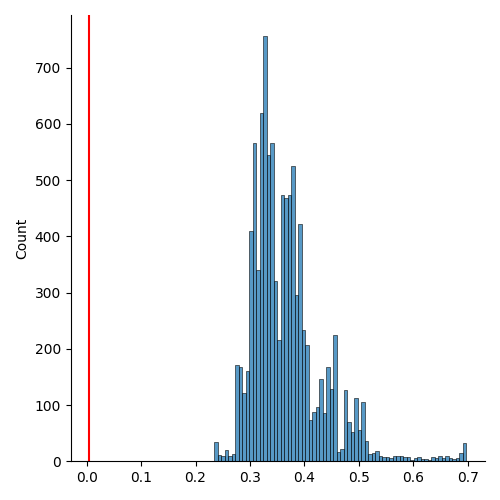}

\includegraphics[width=6.40cm, height=4.80cm]{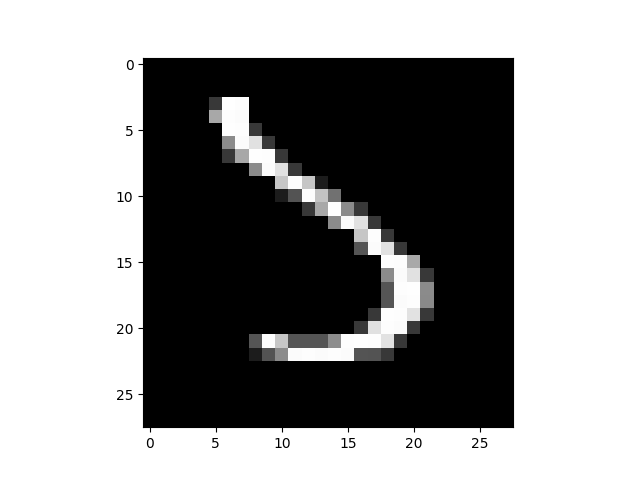}
\includegraphics[width=6.40cm, height=4.80cm]{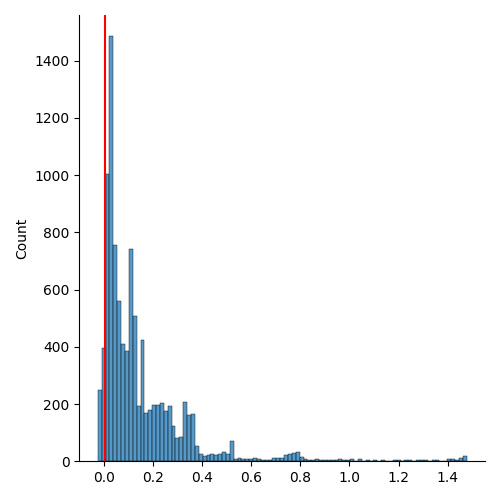}

\includegraphics[width=6.40cm, height=4.80cm]{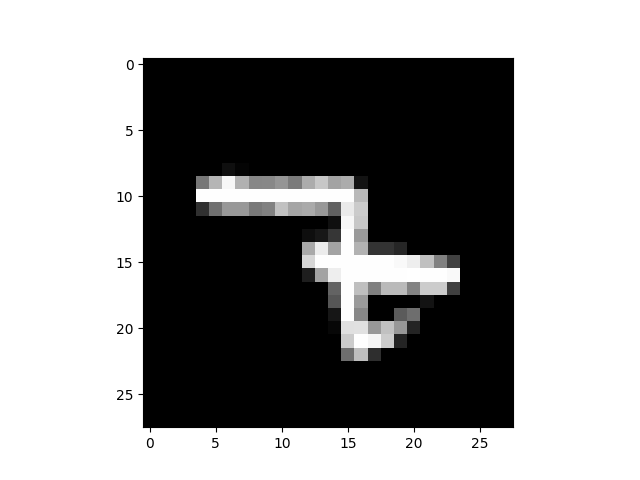}
\includegraphics[width=6.40cm, height=4.80cm]{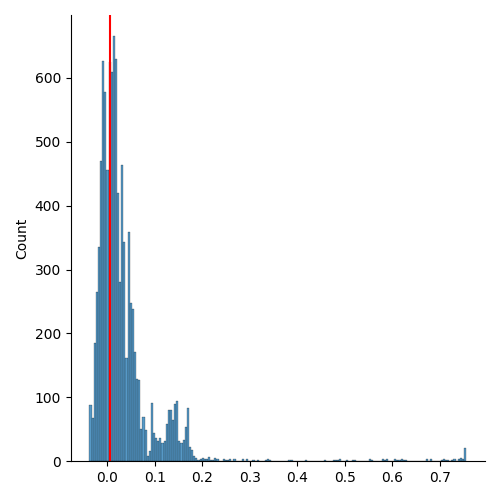}





\includegraphics[width=6.40cm, height=4.80cm]{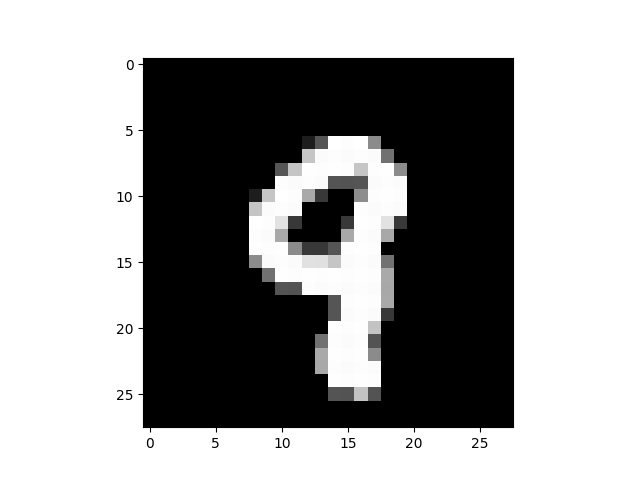}
\includegraphics[width=6.40cm, height=4.80cm]{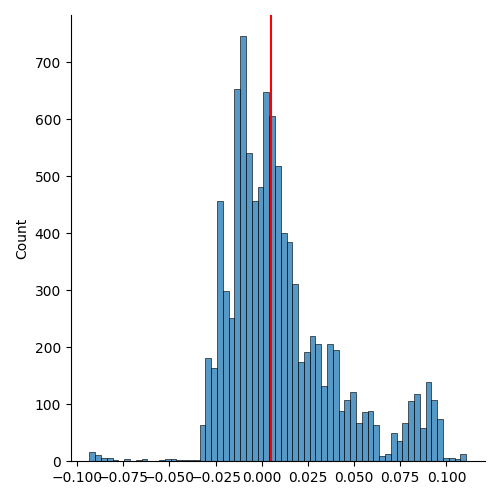}

\includegraphics[width=6.40cm, height=4.80cm]{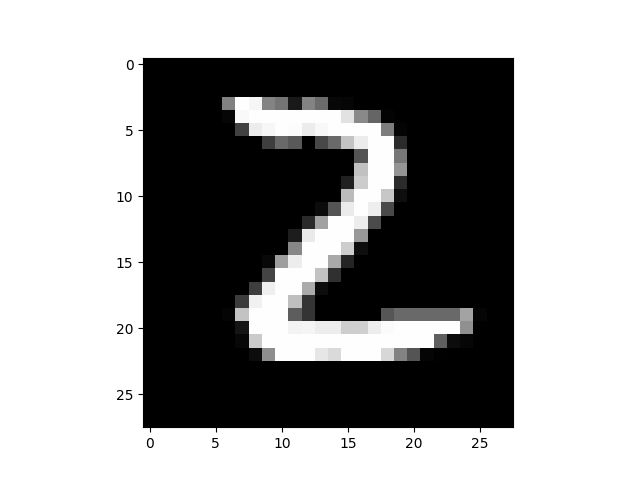}
\includegraphics[width=6.40cm, height=4.80cm]{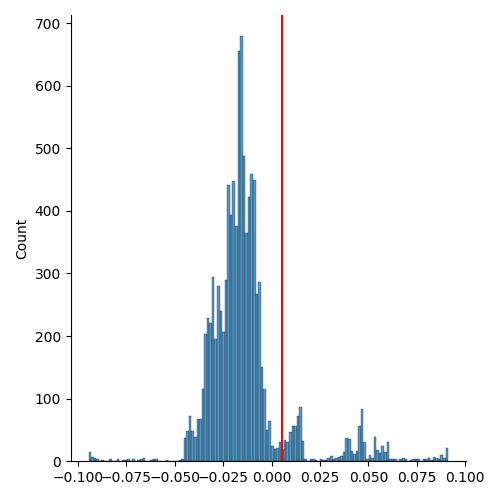}



\includegraphics[width=6.40cm, height=4.80cm]{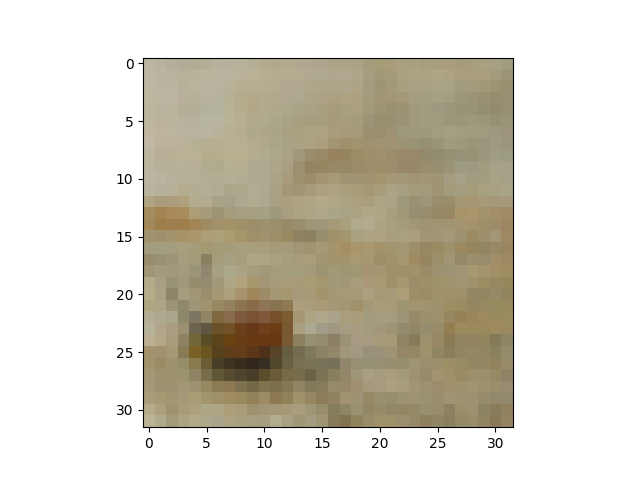}
\includegraphics[width=6.40cm, height=4.80cm]{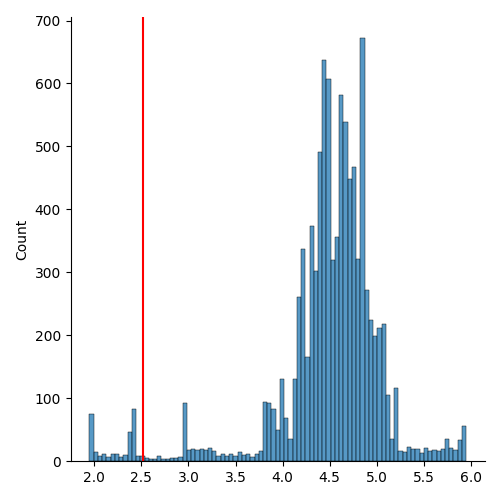}

\includegraphics[width=6.40cm, height=4.80cm]{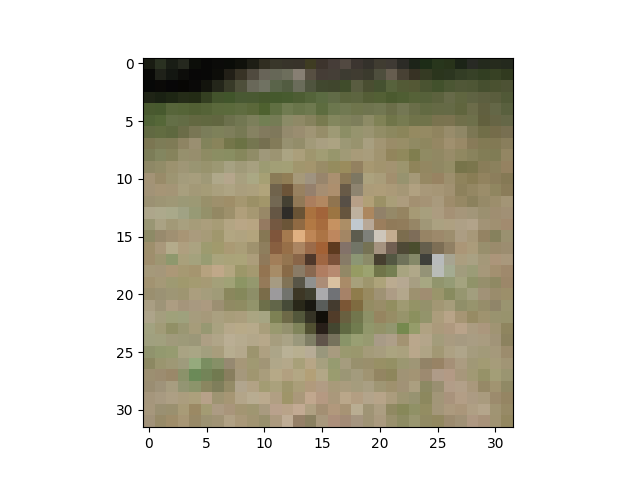}
\includegraphics[width=6.40cm, height=4.80cm]{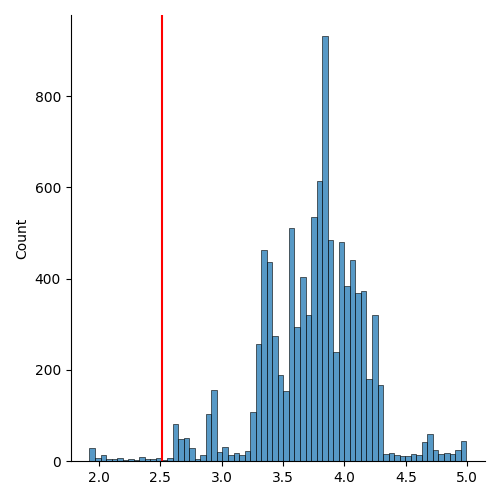}

\includegraphics[width=6.40cm, height=4.80cm]{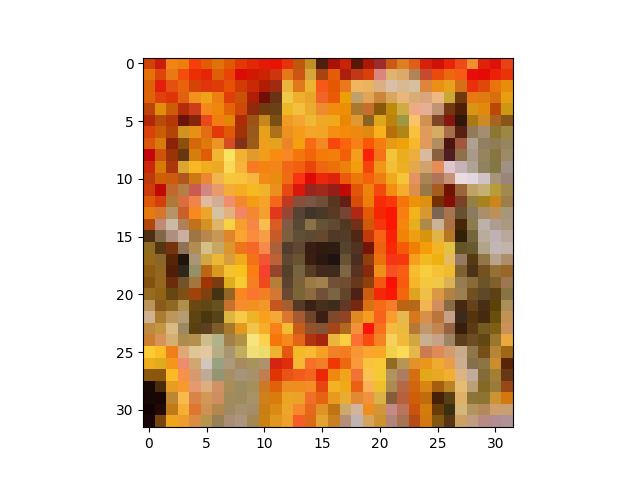}
\includegraphics[width=6.40cm, height=4.80cm]{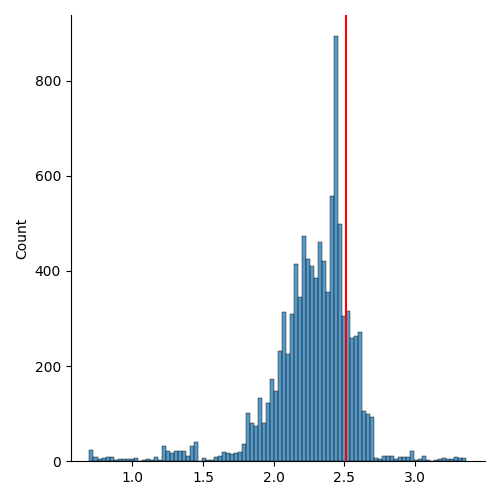}

\includegraphics[width=6.40cm, height=4.80cm]{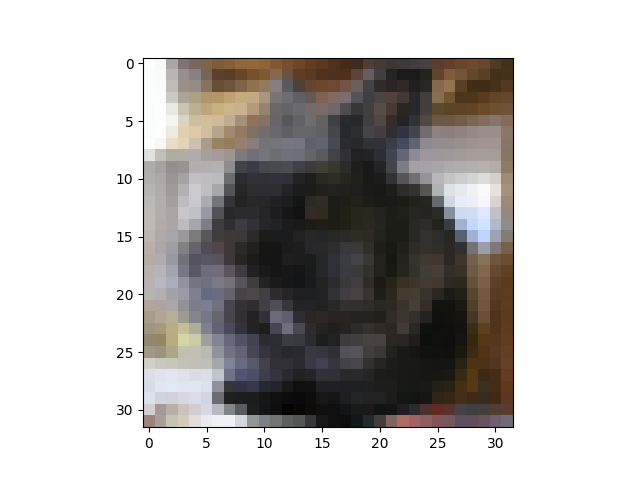}
\includegraphics[width=6.40cm, height=4.80cm]{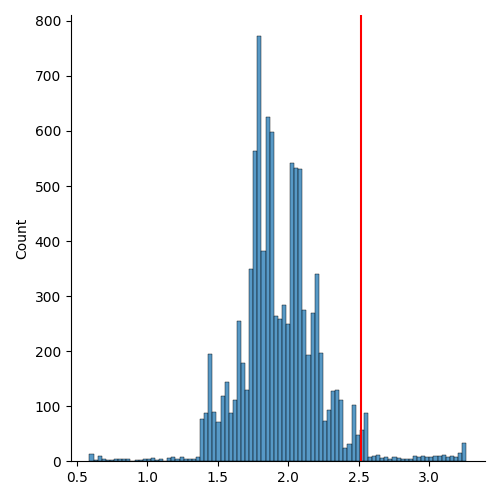}

\includegraphics[width=6.40cm, height=4.80cm]{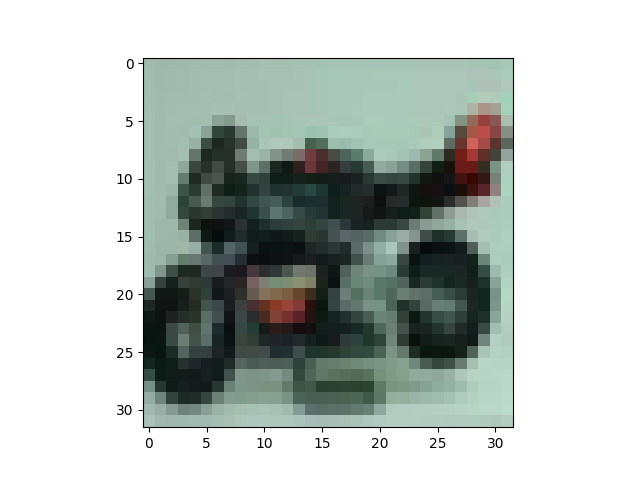}
\includegraphics[width=6.40cm, height=4.80cm]{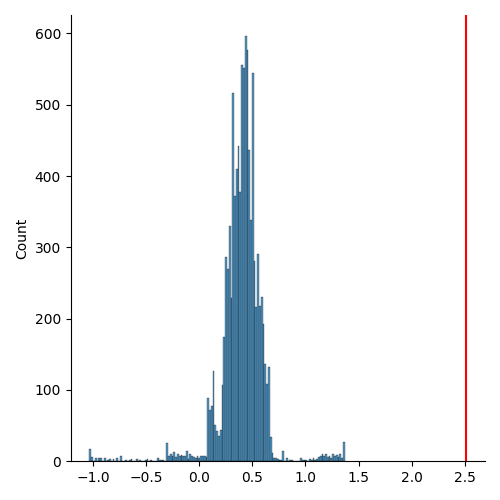}

\section{Future Work}

A promising research direction would be to use this method to filter out bad data for different modalities (such as outliers or incorrect labels). After training the model and IQN, any data with abnormal estimated loss distribution can be reviewed and filtered out. This improves the accuracy compared to simply using the real error from the model. If there's a $10\%$ chance of a defective label, IQN would be able to detect that within its distribution, where it can be flagged for further processing. Future research can also incorporate FQF which is a descendant of IQN that regress a fixed set of taus $\tau$ instead of randomly sampling from the uniform distribution to further improve efficiency by having more fine grained control over the shape of the distribution.

\section{Conclusion}

Our work shows that Implicit Quantile Network is not only beneficial to account for randomness in reinforcement learning settings but can also be used to estimate uncertainty in supervised learning settings. For all the experimented datasets, this method improves accuracy by approximating the distribution of the loss and removing data associated with high loss distributions.

This simple alternative could be crucial for applications where using ensemble learning to train multiple copies of the same model is not computationally feasible. This work have practical applications in medical diagnosis, financial systems and self-driving where predictions that are more likely to have a high error can be disregarded (e.g. for safety concerns or losses in equity). This paper proposes a simple add-on for deep learning models to estimate uncertainty. We advocate that all risk sensitive models adopt this method as an add-on to provide potentially crucial information.

{
\small
\bibliography{neurips_2023}

\begin{thebibliography}{9}
\providecommand{\natexlab}[1]{#1}
\providecommand{\url}[1]{\texttt{#1}}
\expandafter\ifx\csname urlstyle\endcsname\relax
  \providecommand{\doi}[1]{doi: #1}\else
  \providecommand{\doi}{doi: \begingroup \urlstyle{rm}\Url}\fi

\bibitem[Dabney et~al.(2018)Dabney, Ostrovski, Silver, and
  Munos]{dabney2018implicit}
Will Dabney, Georg Ostrovski, David Silver, and R{\'e}mi Munos.
\newblock Implicit quantile networks for distributional reinforcement learning.
\newblock In \emph{International conference on machine learning}, pages
  1096--1105. PMLR, 2018.

\bibitem[Deng(2012)]{deng2012mnist}
Li~Deng.
\newblock The mnist database of handwritten digit images for machine learning
  research [best of the web].
\newblock \emph{IEEE signal processing magazine}, 29\penalty0 (6):\penalty0
  141--142, 2012.

\bibitem[Gal and Ghahramani(2016)]{gal2016dropout}
Yarin Gal and Zoubin Ghahramani.
\newblock Dropout as a bayesian approximation: Representing model uncertainty
  in deep learning.
\newblock In \emph{international conference on machine learning}, pages
  1050--1059. PMLR, 2016.

\bibitem[Huber(1992)]{huber1992robust}
Peter~J Huber.
\newblock Robust estimation of a location parameter.
\newblock In \emph{Breakthroughs in statistics: Methodology and distribution},
  pages 492--518. Springer, 1992.

\bibitem[Krizhevsky et~al.(2009)Krizhevsky, Hinton,
  et~al.]{krizhevsky2009learning}
Alex Krizhevsky, Geoffrey Hinton, et~al.
\newblock Learning multiple layers of features from tiny images.
\newblock 2009.

\bibitem[Mnih et~al.(2013)Mnih, Kavukcuoglu, Silver, Graves, Antonoglou,
  Wierstra, and Riedmiller]{mnih2013playing}
Volodymyr Mnih, Koray Kavukcuoglu, David Silver, Alex Graves, Ioannis
  Antonoglou, Daan Wierstra, and Martin Riedmiller.
\newblock Playing atari with deep reinforcement learning.
\newblock \emph{arXiv preprint arXiv:1312.5602}, 2013.

\bibitem[Ovadia et~al.(2019)Ovadia, Fertig, Ren, Nado, Sculley, Nowozin,
  Dillon, Lakshminarayanan, and Snoek]{ovadia2019can}
Yaniv Ovadia, Emily Fertig, Jie Ren, Zachary Nado, David Sculley, Sebastian
  Nowozin, Joshua Dillon, Balaji Lakshminarayanan, and Jasper Snoek.
\newblock Can you trust your model's uncertainty? evaluating predictive
  uncertainty under dataset shift.
\newblock \emph{Advances in neural information processing systems}, 32, 2019.

\bibitem[Parker(2013)]{parker2013ensemble}
Wendy~S Parker.
\newblock Ensemble modeling, uncertainty and robust predictions.
\newblock \emph{Wiley Interdisciplinary Reviews: Climate Change}, 4\penalty0
  (3):\penalty0 213--223, 2013.

\bibitem[Srivastava et~al.(2014)Srivastava, Hinton, Krizhevsky, Sutskever, and
  Salakhutdinov]{srivastava2014dropout}
Nitish Srivastava, Geoffrey Hinton, Alex Krizhevsky, Ilya Sutskever, and Ruslan
  Salakhutdinov.
\newblock Dropout: a simple way to prevent neural networks from overfitting.
\newblock \emph{The journal of machine learning research}, 15\penalty0
  (1):\penalty0 1929--1958, 2014.

\end{thebibliography}
\bibliographystyle{plainnat}

}


\end{document}